\documentclass{article}

\usepackage{arxiv}

\usepackage[utf8]{inputenc} % allow utf-8 input
\usepackage[T1]{fontenc}    % use 8-bit T1 fonts
\usepackage{hyperref}       % hyperlinks
\usepackage{url}            % simple URL typesetting
\usepackage{booktabs}       % professional-quality tables
\usepackage{amsfonts}       % blackboard math symbols
\usepackage{nicefrac}       % compact symbols for 1/2, etc.
\usepackage{microtype}      % microtypography
\usepackage{lipsum}		% Can be removed after putting your text content
\usepackage{graphicx}
\usepackage{natbib}
\usepackage{doi}
\usepackage{hyperref}
\usepackage{amssymb}
\usepackage{soul, color}
\usepackage{tikz}

\usepackage{amsmath}

\usepackage{algorithm}
\usepackage{algpseudocode}

\usepackage{adjustbox}
\usepackage{rotating}
\usepackage{realboxes}
\usepackage{makecell}
\usepackage{booktabs}
\usepackage{multirow}
\usepackage{diagbox}

\title{REIN-2: Giving Birth to Prepared Reinforcement Learning Agents Using Reinforcement Learning Agents}

\author{ \href{https://orcid.org/0000-0001-9654-9599}{\includegraphics[scale=0.06]{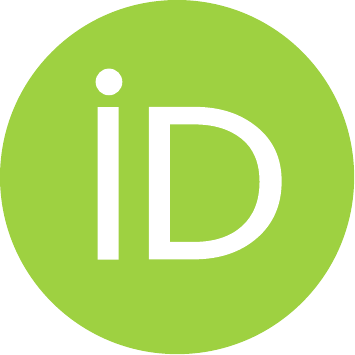}\hspace{1mm}Aristotelis Lazaridis}\thanks{Corresponding author.} \\
	School of Informatics\\
	Aristotle University of Thessaloniki\\
	Thessaloniki, 54124, Greece \\
	\texttt{arislaza@csd.auth.gr} \\
	\And
	\href{https://orcid.org/0000-0003-3477-8825}{\includegraphics[scale=0.06]{orcid.pdf}\hspace{1mm}Ioannis Vlahavas} \\
	School of Informatics\\
	Aristotle University of Thessaloniki\\
	Thessaloniki, 54124, Greece \\
	\texttt{vlahavas@csd.auth.gr} \\
}

\date{}

\renewcommand{\headeright}{}
\renewcommand{\undertitle}{}
\renewcommand{\shorttitle}{REIN-2: Generating RL Agents Using RL}

\hypersetup{
pdftitle={REIN-2: Giving Birth to Prepared Reinforcement Learning Agents Using Reinforcement Learning Agents},
pdfsubject={CS.ML},
pdfauthor={Aristotelis Lazaridis, Ioannis Vlahavas},
pdfkeywords={Deep Reinforcement Learning, Meta-learning, Neural Networks, Games},
}

\begin{document}
\maketitle

\begin{abstract}
Deep Reinforcement Learning (Deep RL) has been in the spotlight for the past few years, due to its remarkable abilities to solve problems which were considered to be practically unsolvable using traditional Machine Learning methods. However, even state-of-the-art Deep RL algorithms have various weaknesses that prevent them from being used extensively within industry applications, with one such major weakness being their sample-inefficiency. In an effort to patch these issues, we integrated a meta-learning technique in order to shift the objective of learning to solve a task into the objective of learning how to learn to solve a task (or a set of tasks), which we empirically show that improves overall stability and performance of Deep RL algorithms. Our model, named REIN-2, is a meta-learning scheme formulated within the RL framework, the goal of which is to develop a meta-RL agent (meta-learner) that learns how to produce other RL agents (inner-learners) that are capable of solving given environments. For this task, we convert the typical interaction of an RL agent with the environment into a new, single environment for the meta-learner to interact with. Compared to traditional state-of-the-art Deep RL algorithms, experimental results show remarkable performance of our model in popular OpenAI Gym environments in terms of scoring and sample efficiency, including the Mountain Car hard-exploration environment.
\end{abstract}

% keywords can be removed
\keywords{Deep Reinforcement Learning \and Meta-learning \and Neural Networks \and Games}

\section{Introduction}
Reinforcement Learning (RL) has been a topic of strong interest in the past, allowing researchers to develop and analyze behaviors that are suitable for solving complex problems. This is due to the reward-or-punish nature of RL, which rewards an agent for performing actions that lead to favorable outcomes, or imposes a punishment when the performed actions diverge from a solution of the problem in hand. In the recent years, Deep Learning methods were incorporated into traditional RL techniques to create the field of Deep Reinforcement Learning (Deep RL). Such methodologies lead to extraordinary results in solving highly complex problems and indicated new, promising directions towards building powerful Artificial General Intelligence systems \citep{lazaridis2020deep}.

However, despite the numerous achievements and applications of Deep RL, there are still critical issues that need to be addressed before these methods are considered “safe” and practical for real-world uses. One such critical problem is the lack of sample efficiency during the training procedure, since even the most sample-efficient state-of-the-art Deep RL models require large amounts of interactions with an environment (i.e. experience) until satisfying performance is achieved, especially when the environment dynamics are highly complex \citep{lazaridis2020deep}. This constitutes a major drawback in cases where the cost of performing wrong actions is high, and thus effective training is essentially impossible.

The attempts to develop sample-efficient Deep RL methods have led to significant progress in the field, with model-based methods being the most promising against this obstacle \citep{deisenroth2011pilco, sutton2018reinforcement}. On the other hand, model-free methods also make use of techniques that target sample-efficiency, such as Prioritized Experience Replay with Importance Sampling \citep{hinton2007recognize, mahmood2014weighted, schaul2015prioritized}, but have not reached yet the fast learning abilities of model-based algorithms.

In this paper, we propose the application of meta-learning in the field of Deep RL for the purpose of creating sample-efficient and high-performance agents out-of-the-box, i.e. Deep RL agents that are already able to solve tasks, through the use of an external Deep RL process. Our algorithm, termed REIN-2 (REINforcement within REINforcement), is in essence a model-free method for training a meta-learning Deep RL model (the \textit{meta-learner}) that, through the use of Deep RL techniques, learns to produce other Deep RL agents (the \textit{inner-learners}) ready to be used within a particular environment. In meta-learning, the goal is to produce an external objective function that optimizes another, internal optimization process; for this reason, meta-learning is also commonly known as the process of “\textit{learning to learn}”.

The proposed system allows for fast and efficient development of Deep RL agents, such as Deep Q-Networks \citep{mnih2015human}, which are not trained with the standard training procedures. The meta-learner has the role of a “factory” that produces trained inner-learners that are ready to solve a given environment, and learns to do so using Deep RL methods. Each produced inner-learner is evaluated on that environment, and its performance is used as feedback to the meta-learner in order to adjust the agent-production process.

In order to implement our methodology, we developed an end-to-end Deep RL framework that allows for easy integration between the meta-learner and the inner-learners, as well as their corresponding Deep RL algorithms. Experimental results show significant abilities of our model to develop high-performance agents on different environments included within the OpenAI Gym platform \citep{brockman2016openai}. More particularly, we achieved to outperform state-of-the-art Deep RL algorithms on toy environments by several orders of magnitude in terms of both performance and number of steps.

The rest of the paper is oragnised as follows. In~\autoref{related} we provide a compact overview of techniques related to meta-learning and Deep RL, highlighting the most notable ones that are similar to ours, along with key differences. In~\autoref{methodology} we describe our methodology thoroughly, including our framework setup, architectural details and further aspects of our implementation. In~\autoref{experiments} we present the experiments conducted using our proposed methodology and their results, along with a direct comparison with state-of-the-art models, and discuss our findings in~\autoref{discussion}. We conclude our work in~\autoref{conclusions}, where we sum up our proposed model and its usefulness in pushing the boundaries of Deep RL, as well as give insights regarding future extensions of the system.

\section{Related Work}
\label{related}
Meta-RL is a field that can be used for optimizing various objectives. For example, meta-RL is commonly used for generalization purposes, i.e. developing a single agent that can succeed in solving multiple problems \citep{gupta2018meta}, as well as for increasing algorithm sample-efficiency \citep{clavera2018learning}. In other words, it can promote adaptation of existing artificial intelligence in new experience by improving an agent’s representation capacity or employing strategies for fast learning using prior knowledge \citep{duan2016rl, Kirsch2020Improving, wang2016learning}. Alternatively, it can be used for learning model parameters (e.g. learning rate) in different tasks \citep{finn2017model}, or allowing agents to communicate efficiently in order to solve sparse rewards problems \citep{parisotto2019concurrent}.

A similar approach has been proposed in Sample-Efficient Automated RL (SEARL) \citep{franke2020sample}, in which the purpose is to generate high-performance and sample-efficient Deep RL agents by using a population-based meta-optimization technique. However, there are two key differences from our proposed model: first, the meta-learning technique in SEARL is based on Neuroevolution \citep{floreano2008neuroevolution, stanley2019designing} and not on Deep RL methods, in contrast to ours, which is a completely Deep RL-based framework, and second, the generated agents are not optimized with respect to their network weights, but to their hyperparameters and network architecture instead.

Likewise, models such as Proximal Distilled Evolutionary Reinforcement Learning (PDERL) \citep{bodnar2020proximal} and Collaborative Evolutionary Reinforcement Learning (CERL) \citep{khadka2019collaborative} are based on the Evolutionary RL framework \citep{khadka2018evolution}, both of which use meta-techniques inspired by the evolutionary biological mechanisms to optimize Deep RL agents. Even though in this kind of evolutionary setting the aim is to discover high-performance Deep RL agents with a reduced sample-efficiency, the meta-learning process itself is not based on Deep RL.

Compared to these meta-techniques, our model uses solely two Deep RL algorithms, used for two different processes that are interconnected. This effectively reduces sample size and improves performance for the task in hand, while incorporating a novel randomized batch vector strategy approach to reduce action space.

\section{Methodology}
\label{methodology}

\begin{figure*}[t]
  \centering
  \includegraphics[width=0.8\textwidth]{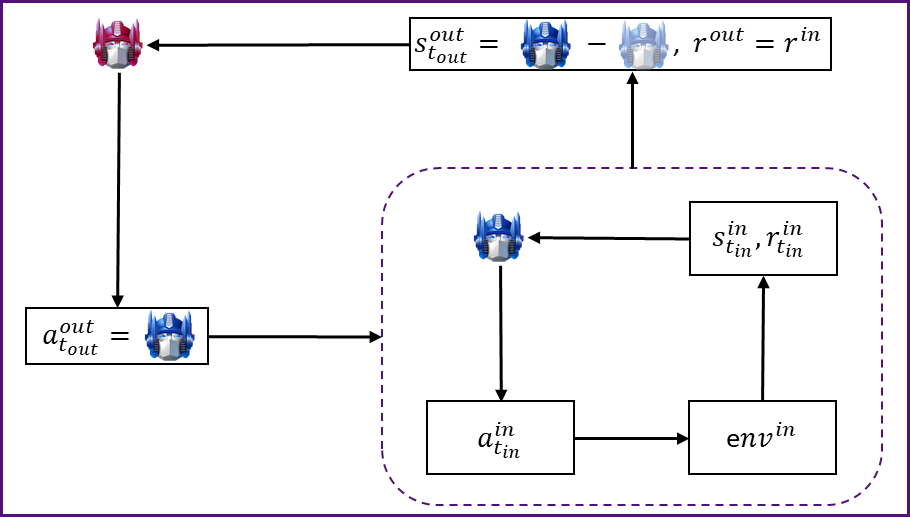}
  \caption{REIN-2 framework pipeline. The meta-learner \includegraphics[scale=0.09, trim={0 12 0 0}]{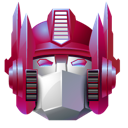} at timestep ${t_{out}}$ produces the inner-learner \includegraphics[scale=0.09, trim={0 12 0 0}]{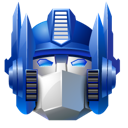}, who is evaluated in the inner environment $en{v^{in}}$. After multiple episodes of interaction with the inner environment, the average reward value ${r^{out}}$ received by the inner-learner during evaluation is used as the reward signal to the meta-learner. Finally, the new state of the outer environment $s_{{t_{out}}}^{out}$ is defined to be the difference between the inner-learners \includegraphics[scale=0.09, trim={0 12 0 0}]{icon_inner_agent.png} and \includegraphics[scale=0.45, trim={0 4 0 0}]{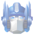} at timesteps ${t_{out}}$ and ${t_{out}-1}$ respectively.}
  \label{diagram}
\end{figure*}

\begin{algorithm*}[t]
\scriptsize
\caption{REIN-2 Pseudocode}\label{algo}
\textbf{Input:} Inner-learner model \({\rm{M}}_\theta ^{in}\)  \\
\hspace*{\algorithmicindent}\hspace*{\algorithmicindent}  Meta-learner model \({\rm{M}}_{\theta '}^{out}\) \\
\hspace*{\algorithmicindent}\hspace*{\algorithmicindent} Inner-environment  \({{\rm{P}}_{{\rm{in}}}}: \left( {{S_{in}},{A_{in}},{R_{in}},{p_{in}}} \right)\) \\
\hspace*{\algorithmicindent}\hspace*{\algorithmicindent} Outer-environment  \({{\rm{P}}_{{\rm{out}}}}: \left( {{S_{out}},{A_{out}},{R_{out}},{p_{out}}} \right)\) \\
\hspace{\algorithmicindent} \textbf{Output:} Trained meta-learner model
\begin{algorithmic}[1]

\State Initialize parameters \({\theta}\) and \({\theta '}\)
\For{\texttt{$t_{out}=1$:$T_{out}$}}
    \State $R_{in} = 0$
    \For{\texttt{$t_{in}=1$:$T_{in}$}}
        \State $R_{in} \leftarrow Evaluate($\({\rm{M}}_\theta ^{in}\), \({{\rm{P}}_{{\rm{in}}}}\)$) + R_{in}$
    \EndFor
    \State $R_{out} \leftarrow \dfrac{R_{in}}{N}$
    \State ${\theta '} \leftarrow $ UpdateMeta(${\rm{M}}_{\theta '}^{out}, R_{out}$)
    \State $a_{t_{out}} \leftarrow GetActionMeta({\rm{M}}_{\theta '}^{out}, s_{t_{out}}), s_{t_{out}}\in {S_{out}}$
     \State ${\theta} \leftarrow a_{t}$
    %\State $w_{1} \leftarrow w_{1}', w_{2} \leftarrow w_{2}'$
\EndFor
\end{algorithmic}
\end{algorithm*}

In order to implement REIN-2, we developed a flexible end-to-end framework for this meta-learning mechanism (Figure~\ref{diagram}). The pseudo algorithm is presented in~\autoref{algo}. More specifically, a Deep RL algorithm has the role of the meta-learner, who learns how to generate suitable inner-learners, i.e. other Deep RL agents who are appropriately tuned to solve given problems (inner environments) without performing any further training procedures of their own. Tuning, in this case, and as explained in detail in the following sections, refers to adjusting a neural network's weights, but can be generalized to replace the trainable target of any learning system. Even though the inner-learners do not use the received rewards from the inner-environments to improve themselves directly, they are still part of the Reinforcement Learning process that takes place within the inner environment, and thus can be considered to be (Deep) RL agents.

The Deep RL algorithms of the meta-learner and the inner-learners, as well as the inner environments, are used as input arguments to our framework in a way that is described below, with a minimum amount of adjustments. This framework makes use of the stable-baselines library \cite{hill2018stable} for executing implemented versions of algorithms.

\subsection{Problem Formulation}
\label{problem}
The inner environment \({{\rm{P}}_{{\rm{in}}}}\) that we wish to solve is formulated as a Markov Decision Process (MDP), defined as the tuple \(\left( {{S_{in}},{A_{in}},{R_{in}},{p_{in}}} \right)\), where \({S_{in}}, {A_{in}}\) and \({R_{in}}\) define the state space, action space and reward function of the problem respectively, while \({p_{in}}\) denotes the probability transition function.

Additionally, let \({{{\rm{M}}^{in}}\left( {{\theta}} \right)}\) be an arbitrary Reinforcement Learning model for the role of the inner-learner, where \({\theta} \in {{{W}}^k}\) denotes the model's parameters, and \({{{W}}^k}\) is a \(k\)-dimensional space of values for the parameters. To be more compact, we shall denote the inner-learner as \({\rm{M}}_\theta ^{in}\).

Thus, the optimization objective \({J_{in}}\) for the particular problem of solving \({{\rm{P}}_{{\rm{in}}}}\) using \({\rm{M}}_\theta ^{in}\) can be defined as:
\begin{equation}
\mathop {\arg \max }\limits_{{\theta}} {J_{in}}\left( {\rm{M}}_\theta ^{in},{{\rm{P}}_{{\rm{in}}}} \right)
\end{equation}
%\[\textrm{s.t.} \quad {\theta} \in {{{W}}^k}\]

Moreover, let \({{\rm{P}}_{{\rm{out}}}}\) denote the outer environment with which the meta-learner interacts. \({{\rm{P}}_{{\rm{out}}}}\) is defined as an MDP formed by the tuple \(\left( {{S_{out}},{A_{out}},{R_{out}},{p_{out}}} \right)\), where \({S_{out}} = {W^k}\), \({A_{out}} = {W^k}\) and \({R_{out}} = {R_{in}}\) are the problem's state space, action space and reward function respectively, while \({p_{out}}\) denotes the probability transition function.

The meta-learner is a Reinforcement Learning model \({{{\rm{M}}^{out}}\left( {{\theta '}} \right)}\), where \({\theta '} \in {{{H}}^l}\) denotes the model's parameters, with \({{{H}}^l}\) being an \(l\)-dimensional space of values for the parameters. Similarly as before, to be more compact, we shall denote the meta-learner as \({\rm{M}}_{\theta '}^{out}\).

Through a function \(f\), the meta-learner picks an action \({a_{out}} \in {A_{out}}\) and executes it in the environment, i.e. \(f\left( {{\rm{M}}_{\theta '}^{out},{{\rm{P}}_{{\rm{out}}}}} \right) = {a_{out}}\). However, since \({A_{out}} = {W^k}\), then \({a_{out}} \in {W^k}\), and thus the action can be used in place of \({\theta}\) in the original optimization problem:
\begin{equation}
\begin{aligned}
\mathop {\arg \max }\limits_{{\theta}} {J_{in}}\left( {{\rm{M}}_\theta ^{in},{{\rm{P}}_{{\rm{in}}}}} \right) =  
\mathop {\arg \max }\limits_{{a_{out}}} {J_{in}}\left( {{\rm{M}}_{a_{out}} ^{in},{{\rm{P}}_{{\rm{in}}}}} \right) = \\
= \mathop {\arg \max }\limits_{{\theta '}} {J_{in}}\left( {{{\rm{M}}^{in}}\left( {{{f\left( {{\rm{M}}_{\theta '}^{out},{{\rm{P}}_{{\rm{out}}}}} \right)}}} \right),{{\rm{P}}_{{\rm{in}}}}} \right)
\end{aligned}
\end{equation}
%\[\textrm{s.t.} \quad {\theta'} \in {{{W}}^k}\]

Therefore, the optimization objective shifts to finding the optimal parameters \(\theta'\) of the meta-learner. With these parameters, the meta-learner can produce the actions that correspond to optimal parameters \(\theta\) of the inner-learner, subsequently solving the inner environment.

%\[\textrm{s.t.} \quad {\theta'} \in {{{W}}^k}\]

\subsection{REIN-2 Framework}
\label{rein2framework}

A top-level layer is created for the REIN-2 system, which is the input layer that accepts the Deep RL algorithms (i.e. the meta-learner and inner-learner algorithms that will be used) and the inner environment to be solved.

Then, a conversion layer receives this information and processes the contents of the inner algorithms and inner environment appropriately. This is a critical process, since a custom outer environment is created to formulate another RL problem. The outer environment consists of the useful information that occurs during the interaction between the inner-learner with the inner environment. The conversion layer implements an action space and a state space, as well as a reward function, all of which define the RL problem that the meta-learner attempts to learn to solve by interacting with the outer environment.

The action space defined by the outer environment depends on the type of the inner-learner. As described previously, the main goal is to train a meta-learner that can generate other agents that are capable of solving particular environments. This ultimately means that the outer environment defines an action space that is equivalent to an inner learner, or, in other words, the parameter space of the algorithm used to solve the inner environment. 

In order for the meta-learner to become efficient at producing capable inner learners for the given environments, the state space of the outer environment has to provide meaningful and useful information regarding the interaction that occurs between the inner learner and the inner environment. After experimenting with various state spaces, we settled for the scenario where a state of the outer environment is defined as the difference between two consecutive inner learners. 

The reward function of this system is also a highly sensitive parameter that has a direct effect on the meta-learner’s performance in producing optimized inner-learners that are capable of solving the given environment. In our experiments, we defined it to be the average reward of the inner-learner when evaluated in multiple episodes within the inner environment. This choice guarantees that there is no information exchange between the meta-learner and the inner-learners other than what the inner-learner is handed by the inner environment. The averaging is also important since it allows the meta-learner to get an accurate view of the resulting agent’s performance, but a smaller evaluation window for even smaller wall-clock times during execution can be applied, trading off performance estimation accuracy.

\subsection{Characteristics of Learners}
\label{characteristics}

The state space of the outer environment was defined as ${\bf{S}} \in \mathbb{R}{^k}$, where $k$ is the number of weights of the inner-learner’s Deep Q-Network. Similarly, the action space was defined as ${\bf{A}} \in \mathbb{R}{^k}$. At timestep $t$, the action vector ${{\bf{a}}_t}$ corresponds to the inner learner’s DQN weight values, which remain frozen throughout the whole interaction process with the inner environment or, intuitively, it corresponds to a generated agent that will be evaluated within the inner environment at that timestep. Consequently, at timestep $t$, the state vector ${{\bf{s}}_t}$ is defined as ${{\bf{s}}_t} = {{\bf{a}}_t} - {{\bf{a}}_{t - 1}}$, which is essentially the difference in network weight values between the last inner agent and the current inner agent.

Each timestep for the meta-learner corresponds to $N$ episode runs of the inner environment, which allows us to estimate accurately the performance of the generated agent based on his average reward during these runs. This average reward is used as the reward signal for the meta-learner.

The intuition behind these choices corresponds to the assumption that useful information lies within the differences between the meta-learner’s choices. Since the meta-learner generates different agents at each timestep that perform differently on the same inner environment, it should be able to pick up on significant weights of the agents’ network. Subsequently, the meta-learner can learn to ignore specific traits of the agent that are of small importance with regard to the specific task in hand, and boost other, useful skills.

\subsection{Randomized Batch Vector Strategy}
\label{rbv}

The single goal for the meta-learner is to adjust the elements of the action vector ${{\bf{a}}_t}$ in order for the inner-learner (e.g. Q-Network) to solve the environment without any training. However, this vector has a relatively huge length, consisting of several thousands of parameters, which makes the learning process for the meta-learner even more difficult, and sometimes impractical due to constraints posed by computational resources. For this reason, we proposed a Randomized Batch Vector (RBV) strategy: we use a random subset (batch) of specified size from the action vector, and let the meta-learner focus on adjusting the weights included only within that batch. This subset is fixed throughout the whole training process (i.e. the same network weights are tuned) while the rest of the weights remain untouched, with their values being the values set during initialization of the network throughout the whole training process.

The RBV strategy that we used is one approach for defining a smaller state space and achieving high performance. REIN-2 can be potentially modified in order to use a different implementation for this purpose.

The purpose of this strategy is to restrict the number of parameters that the model has to learn. It is only a primal approach to tackle the problem of dealing with large state spaces, and even though there is no guarantee that the important weights are selected, it proved to yield a significant performance boost in our model. Since our goal is to provide a novel methodology to the research community that has various configurable modules, extending REIN-2 to use other methods than random sampling with RBV, is an interesting case for future work.

For the experiments against state-of-the-art models we used an RBV percentage of 1\%, but we also report performance comparison between our models using different RBV values.

\section{Experimental Results}
\label{experiments}

We applied our methodology in various scenarios, to evaluate both its efficiency and efficacy. As described in detail below, the results are promising and a scaled-up version of our system can prove to be very useful towards building a powerful, generalized end-to-end Deep RL system.

The analysis of the proposed methodology was performed using the CartPole-v1, Acrobot-v1 and MountainCar-v0 \cite{barto1983neuronlike, geramifard2015rlpy, moore1990efficient} as the inner environments, A2C \cite{bhatnagar2009natural, mnih2016asynchronous} and PPO \cite{schulman2017proximal} algorithms for the meta-learner, and DQN \cite{mnih2015human} for the inner-learner.

For our experiments, we measured the performance of a meta-learner separately for each inner environment, defined as the average score of the inner learner in $N=10$ evaluation episodes, used a size of $1\%$ for the RBV strategy, and compared it against the performance of state-of-the-art algorithms.

\begin{figure*}[t]
  \centering
  \includegraphics[width=1\textwidth]{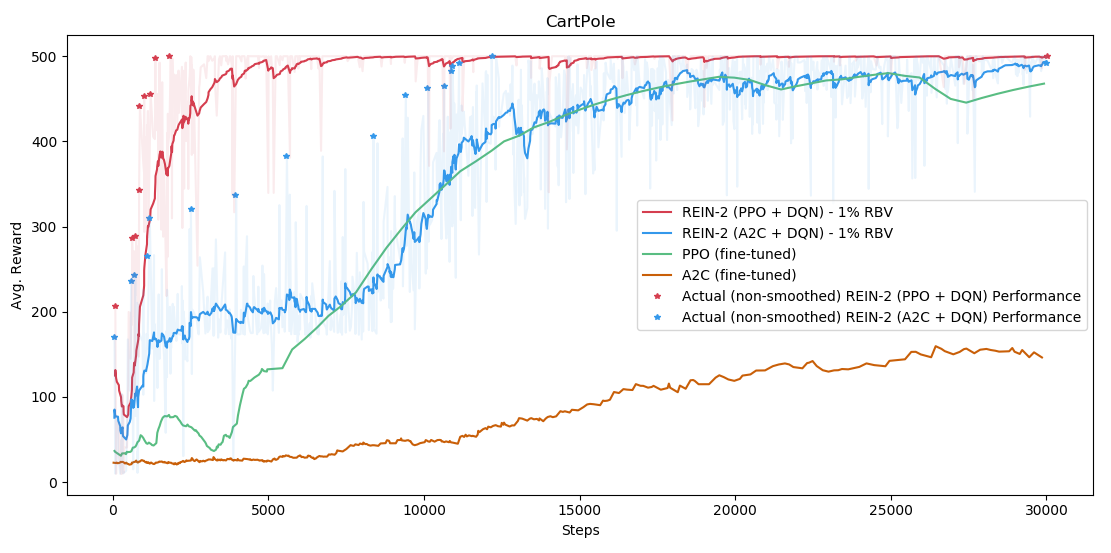}
  \caption{REIN-2, compared to fine-tuned versions of A2C and PPO, generates high-performance agents early within the training procedure for the CartPole environment. It identifies the necessary behavior that the spawned agents should have to solve the problem (i.e. achieve score of 500), and begins producing them more consistently during training. Shaded areas indicated the non-smoothed performance of our models, while stars point to some of these scores.}
  \label{cartpole}
\end{figure*}

\subsection{Environments}

For each environment that we used to test our model, we performed a loose fine-tuning procedure and compared the average performance per timestep over 3 different seeds against A2C and PPO, which were carefully fine-tuned specifically for each environment. Those hyperparameters were obtained from the RL Baselines Zoo project \cite{rl-zoo}. For the experiments, we used an RBV size of only $1\%$ of the original network.

In the CartPole-v1 environment, the goal is to keep a pendulum upright and prevent it from falling, for as long as possible. A reward of +1 is given to the agent for every timestep that the pole remains upright. The episode ends when a particular number of steps (500) has been reached, and, considering that the starting position of the pole is always upright, maximum reward is 500. Performing better than a random agent in CartPole is relatively easy, but achieving maximal performance is harder to achieve \cite{hill2018stable}.

Our model's performance in CartPole can be seen on Figure~\ref{cartpole}. Essentially, each timestep depicts the smoothed average performance of a different agent, and as it can be seen, both PPO and A2C as the meta-learners for REIN-2 produce agents that perform relatively well from the beginning of the learning process. For the rest of the environments, high-performance agents were generated from the very first steps, outperforming the rewards reached by state-of-the-art models after several thousands of steps.

The model’s capability to learn how to generate strong agents are also evident in the experiments. Especially in the case of PPO as the meta-learner, it learned to generate consistently better agents in only a small fraction of the steps required by the fine-tuned state-of-the-art methods A2C and PPO to learn how to receive high rewards in this environment.

\begin{figure*}[t]
  \centering
  \includegraphics[width=1\textwidth]{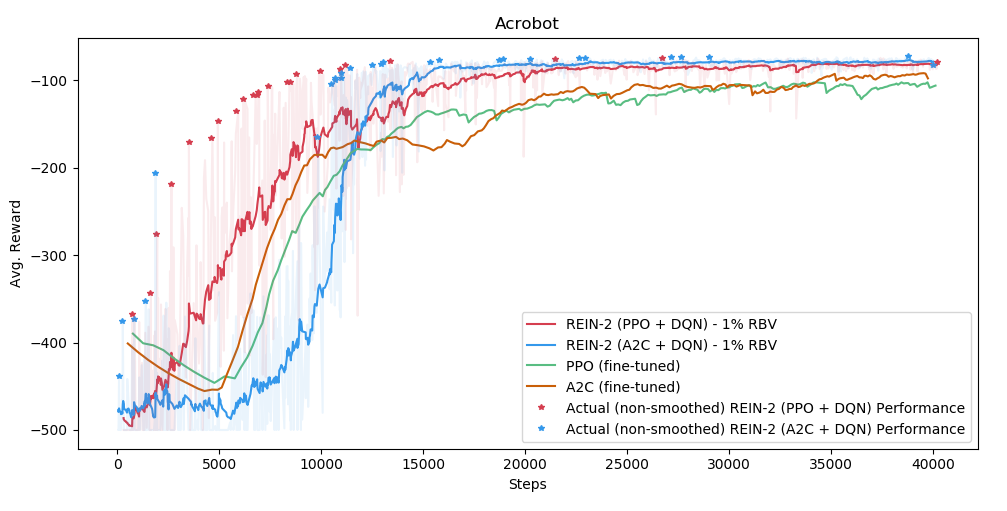}
  \caption{REIN-2, compared to fine-tuned versions of A2C and PPO, generates high-performance agents early within the training procedure for the CartPole environment. It identifies the necessary behavior that the spawned agents should have to solve the problem, and begins producing them more consistently during training. Shaded areas indicated the non-smoothed performance of our models, while stars point to some of these scores.}
  \label{acrobot}
\end{figure*}

Then, we proceeded into experimenting with a more complex toy environment, the Acrobot-v1. In this environment, the task is to swing a two-link two-joint object in a way that it reaches a horizontal base line. The values of the reward function in this case are always negative, indicating that the optimization problem defined requires minimizing the number of steps until an episode ends. More particularly, the agent receives a reward of -1 for each timestep, and the termination condition is met either when target height is reached, or 500 timesteps have passed. Similarly to CartPole, our model performs equally well in managing to find suitable inner agents that can solve this task very soon in the beginning of the training process (Figure~\ref{acrobot}). Additionally, the meta-learner learns to pick up the key elements required to produce better agents more and more frequently.

\begin{figure*}[t]
  \centering
  \includegraphics[width=1\textwidth]{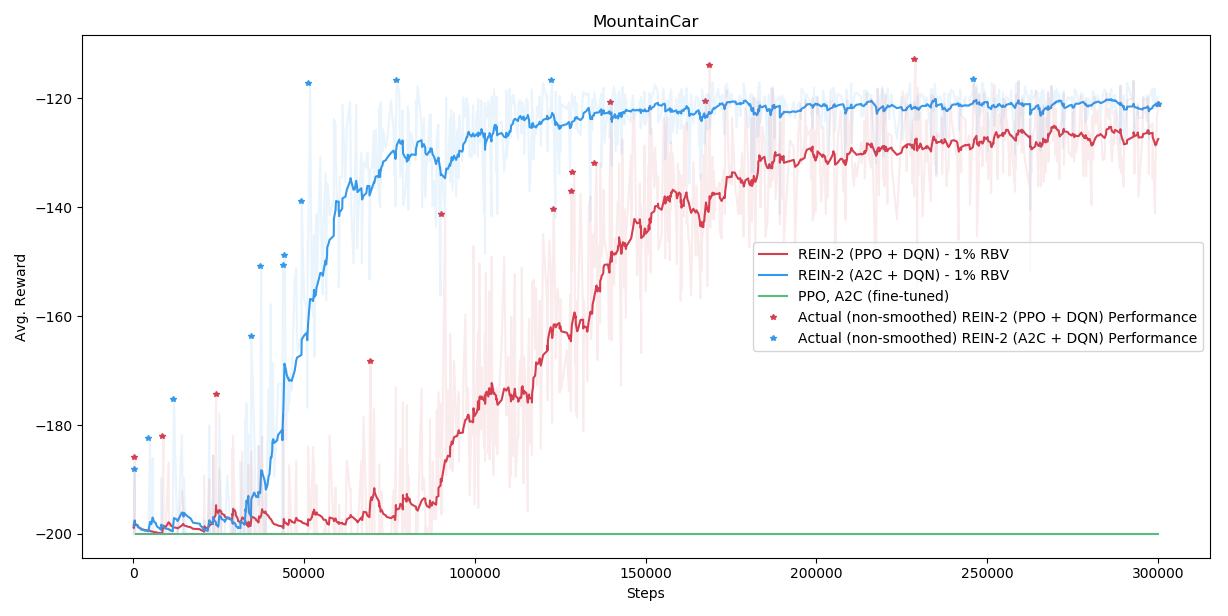}
  \caption{In contrast to PPO and A2C, REIN-2 can reach the goal in the MountainCar problem multiple times from the beginning of the training process. Shaded areas indicated the non-smoothed performance of our models, while stars point to some of these scores.}
  \label{mountaincar}
\end{figure*}

The last experiment was performed in the well-known MountainCar environment, which is commonly used for evaluation of RL agents and mainly their exploration capabilities, as it is a sparse reward problem, on top of the other complexities of the environment dynamics. Even state-of-the-art Deep RL algorithms struggle achieving decent performance on this task, unless equipped with very special exploration abilities. Similarly to the Acrobot environment, the agent receives a reward of -1 for each timestep, and the termination condition is met when the agent reaches the top of the hill (top = 0.5 position). The problem is considered to be solved when the learned policy manages to achieve a mean reward of -110 over 100 consecutive runs.

Despite these difficulties that are present in the MountainCar environment, REIN-2 proved to be able to produce agents that can reach the goal of this task fairly soon after the training procedure starts. In fact, as it can be seen from Figure~\ref{mountaincar}, the meta-learner can produce various agents that perform near-optimally in a relatively few timesteps, whereas PPO and A2C never manage to solve the problem in a satisfying timeframe.

For the purposes of evaluating the learning abilities of the meta-learner, we allowed the experiment to run for several thousand steps. Our observations were that performance of the produced inner-agents kept improving and converging to a near-optimal solution behavior during training, and production frequency of these high-performance agents by the meta-learner was increasing as well.

For a more comparable view of our model's performance, in~\autoref{meta_table} we report the scores achieved in the aforementioned in the experiments at various timesteps throughout the training procedure (i.e. snapshots). It is evident that both high performance and sample efficiency are achieved using our meta-learning methodology, outperforming the fine-tuned state-of-the-art models PPO and A2C in all environments.

\renewcommand\theadalign{bl}
\begin{table}[!t]
  \centering
\caption{Snapshots of our model's average performance scores against PPO and A2C, at specific timesteps. Cell values indicate the score of each model in the respective environment, averaged over 3 different seeds. Bold values indicate the highest (i.e. best) score of the column. In all environments, and at various timesteps, our model's variants outperformed state-of-the-art algorithms PPO and A2C. In most cases, the meta-learner produced inner-agents that were capable of achieving near-optimal scores very soon in the training process.}  
\resizebox*{\textwidth}{!}{% Table generated by Excel2LaTeX from sheet 'Sheet1'
    \begin{tabular}{|c|c|c|c|c|c|c|c|c|c|}
    \hline
          & \multicolumn{3}{c|}{\textbf{CartPole-v1}} & \multicolumn{3}{c|}{\textbf{Acrobot-v1}} & \multicolumn{3}{c|}{\textbf{MountainCar}} \\
          \hline
    \diagbox{Algorithm}{Step}
& \textbf{75} & \textbf{1000} & \textbf{2500} & \textbf{700} & \textbf{3500} & \textbf{10000} & \textbf{250} & \textbf{70000} & \textbf{150000} \\
    \hline
    \textbf{PPO} & 22.45 & 30.76 & 49.64 & -500  & -500  & -208.17 & -200  & -200  & -200 \\
    \textbf{A2C} & 24.82 & 17.33 & 53.56 & -500  & -500  & -216.39 & -200  & -200  & -200 \\
    \textbf{REIN-2 (PPO + DQN)} & \textbf{207.2} & \textbf{439.68} & \textbf{466.96} & \textbf{-425.43} & \textbf{-257.76} & \textbf{-94.05} & \textbf{-186.03} & -177.64 & -140.87 \\
    \textbf{REIN-2 (A2C + DQN)} & 66.76 & 146.86 & 203.85 & -454.77 & -500  & -344.33 & -188.27 & \textbf{-127.52} & \textbf{-122.37} \\
    \hline
    \end{tabular}%
}
\label{meta_table}
\end{table}

\begin{figure*}[t]
  \centering
  \includegraphics[width=1\textwidth]{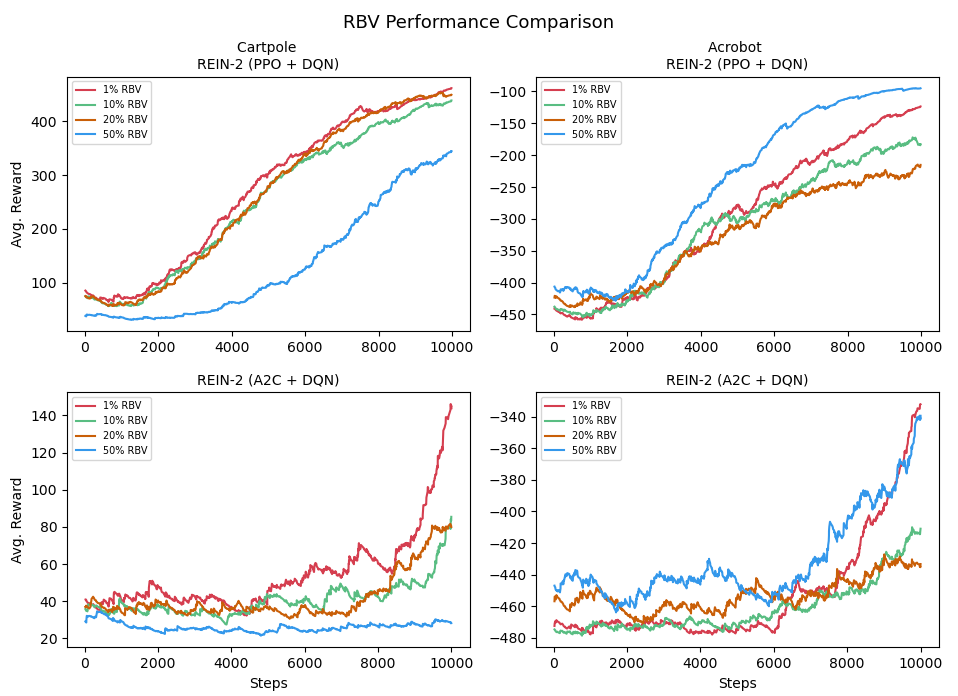}
  \caption{Performance scores of the REIN-2 models with different RBV values, in Cartpole and Acrobot.}
  \label{figure_rbv}
\end{figure*}

Additionally, we performed experiments with various RBV percentages in the Cartpole and Acrobot environments, the results of which are presented in~\autoref{figure_rbv}. Setting RBV to 100\% is equivalent to using the whole set of network weights of an inner-learner to form the state space, which becomes enormous, since each state has a length in the scale of hundreds of thousands, or millions. Therefore, due to computational constraints, we present results for percentages of up to 50\%. For the Cartpole task, lower RBV values have similar effect on the model and allow it to achieve high performance, while 50\% has a negative effect on performance. This implies that the meta-learner learns about the state space at a slower pace since it is larger and more difficult to explore efficiently, but the task's dynamics are far simpler and thus can be solved using a smaller state space. It should be noted however that the models still manage to learn how to improve at generating inner-learners that perform well.

On the other hand, the 50\% RBV in the Acrobot task seems to achieve highest performance compared to the other models, while the 1\%, 10\% and 20\% models have a slightly worse performance but they all have a similar convergence rate. This indicates that Acrobot, which is a more complex task, requires the meta-learner to need more information from the stata space in order to learn the required behaviors of the inner-learners in order to perform well. Even though the meta-learner has more difficulty ``becoming familiar” with the state space of the outer-environment, this trade-off in information that it receives is to its advantage.

\section{Discussion}
\label{discussion}

In all our experiments, REIN-2 showed significant abilities to outperform surpass current state-of-the-art Deep RL algorithms in terms of both performance and convergence speed, always maintaining a sample-efficient behavior. Even in the MountainCar environment, a well-known hard-exploration problem that many state-of-the-art Deep RL methods fail to solve, our model achieved to find agents that were able to reach the goal relatively easy. Our assumptions regarding the formulation of the outer RL process, such as the definitions of the state/action space and the reward function, proved to be clear enough for the meta-learner to achieve its purpose, which is to learn how to spawn capable agents.

It is noteworthy that our method beholds two distinguishing properties: the ability to generate high-performance agents from the very early stages of the training procedure, and the ability to learn how to adapt to the inner-environments, generating consistently better agents. Even though the latter property is usually to be expected in a typical RL problem, the former one is extremely rare.

To this end, our Randomized Batch Vector strategy proved to be very beneficial for achieving these results, and at the same time indicates that significantly smaller networks can be used to solve seemingly difficult problems. Setting RBV percentage to higher values and therefore using a larger portion of the full state space is beneficial when the given problem has higher complexity, and harms model performance when the problem is simpler. This is expected because the length of a state vector, depending on the environment, can reach the order of millions, therefore a balance between state space size and the necessary information received by the meta-learner is sufficient. The struggle of learning in large and complex state spaces is present in any Deep Reinforcement Learning agent, therefore, using a component for reducing the full state space, in this case RBV strategy, is crucial to REIN-2. Our goal is to provide a novel methodology, which has various configurable modules, and the RBV strategy is one implementation for the module responsible for defining a smaller state space and achieving high performance.

Even though experimental results indicate that the RBV strategy is sufficient for REIN-2 to outperform state-of-the-art Deep RL algorithms in the evaluation testbeds, we believe that in-depth research on these strategies as part of our proposed system can further improve its overall performance.

With REIN-2 we propose a new aspect in which a Reinforcement Learning process can be redefined using a meta-learning approach, forming a new Reinforcement Learning problem (i.e. the problem defined by the outer environment). This way, by being able to encapsulate the given (inner) environment in a custom (outer) environment, one can configure REIN-2 in many ways and optimize it accordingly, such as by setting the RBV (or another) strategy for reducing the complexity of the state space when defining a state space that corresponds to the network architecture of the inner-learner, or defining a different state space that is more compact and informative, as well as defining a different reward function for the outer environment that is different from the reward function in the inner-environment. The advantage is that all these parameters of the REIN-2 system are configurable and result in different levels of performance, while the original problem remains untouched.

\section{Conclusions and Future Work}
\label{conclusions}

Deep Reinforcement Learning has various limitations that need to be addressed before becoming an industry standard when it comes to AI applications. Even state-of-the-art algorithms require large amounts of experience to solve complex problems and often reach suboptimal and unsatisfying solutions. In this paper we propose REIN-2, a novel end-to-end Deep RL framework that targets these limitations, by designing a meta-learning scheme that uses a pair of Deep RL algorithms. We formulated a new RL problem in which the meta-learner is trained to generate inner-learners capable of solving a given task in a highly sample-efficient way, by tuning properly their network weights. For this purpose, and as a technique to reduce the large action space that occurred in our setup, we also applied a Randomized Batch Vector strategy that limits the meta-learner’s operations to only a subset of the inner-learner’s network.

The concept of our model can be further expanded by introducing multiple meta-learners controlled by another, higher-level meta-learner, as in a 3-tier architecture. In the same manner, introducing even more layers of learners will form a network, where each node is a learner generated by higher-level learners. Equivalently to the training procedure of neural networks with backpropagation, this network will be trained using RL methods. The ultimate purpose of this conceptualized model is to be able to generalize in novel, unrelated environments, by allowing different inner-learners solve different tasks within the same network of learners, passing information along to higher-level meta-learners, subsequently training them to produce capable agents regardless of the environment. This can be seen as a meta-Reinforcement Learning “factory” for the solution of a diversity of tasks. Implementing and assessing this idea of a system is part of the authors’ future work.

\bibliographystyle{unsrtnat}
\bibliography{REIN-2}  %%% Uncomment this line and comment out the ``thebibliography'' section below to use the external .bib file (using bibtex) .





% \end{thebibliography}

\end{document}